\documentclass[10pt,twocolumn,letterpaper]{article}

\usepackage[pagenumbers]{cvpr} % To force page numbers, e.g. for an arXiv version
\usepackage{soul}
\usepackage{multirow}
\usepackage{tikz}
\usepackage{bbding}
\usepackage{xcolor}
\usepackage{float}

\newcommand{\secref}[1]{Sec.~\ref{#1}}
\newcommand{\figref}[1]{Fig.~\ref{#1}}
\newcommand{\tabref}[1]{Tab.~\ref{#1}}

\newcommand{\gh}[1]{{\color{black} {#1}}}
\newcommand{\fw}[1]{{\color{black} {#1}}}
\newcommand{\chl}[1]{{\color{black} {#1}}}

\definecolor{cvprblue}{rgb}{0.21,0.49,0.74}
\usepackage[pagebackref,breaklinks,colorlinks,allcolors=cvprblue]{hyperref}

\def\paperID{8254} % *** Enter the Paper ID here
\def\confName{CVPR}
\def\confYear{2025}
\usepackage{orcidlink}
\title{ChatSplat: 3D Conversational Gaussian Splatting}

\author{%
    Hanlin Chen$^{1}$\!\!\quad Fangyin Wei$^{2}$\!\!\quad Gim Hee Lee$^1$ \\
    $^1$\, National University of Singapore % School of Computing, 
    $^2$\, Princeton University \\ 
    \texttt{hanlin.chen@u.nus.edu} \quad \texttt{gimhee.lee@nus.edu.sg} \\
    {\tt \href{https://github.com/HLinChen/ChatSplat}{\textbf{https://github.com/HLinChen/ChatSplat}}}
}

\begin{document}

\twocolumn[{%
\renewcommand\twocolumn[1][]{#1}%
\maketitle
\vspace{1.5cm}
\vspace{-3\baselineskip}
\vspace{-3\baselineskip}
\begin{center}
\centering
\setlength{\tabcolsep}{0.5pt}
\captionsetup{type=figure}
{\footnotesize
\renewcommand{\arraystretch}{0.5} 

\begin{tikzpicture}
\node (img) {
\begin{tabular}{c c c c c c}
 \includegraphics[width=1\linewidth]{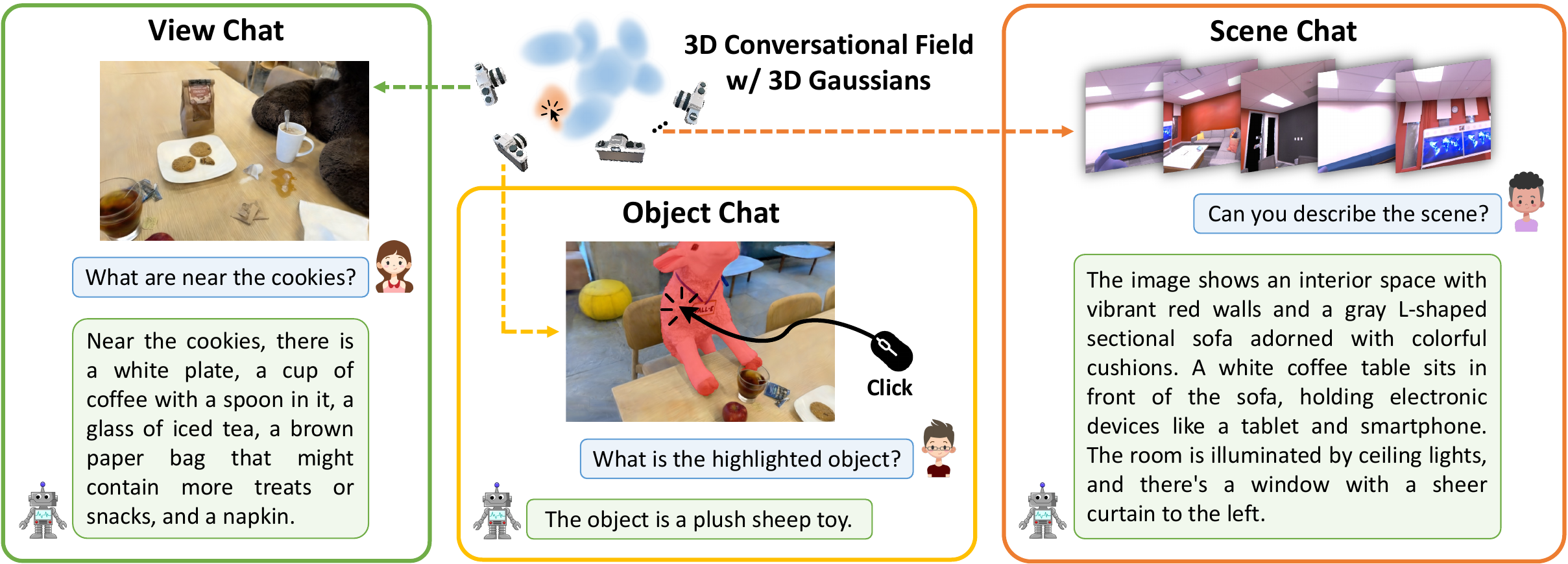}
\end{tabular}
};
\end{tikzpicture}

}
\vspace{-5mm}
\vspace{-0.5\baselineskip}
\hspace{20pt}\captionof{figure}{\label{fig:teaser}ChatSplat is the first 3D Gaussian Splatting-based approach that enables conversational interaction with a 3D environment across multiple levels, including view, object, and the entire scene. The key idea is to learn a 3D conversational field with 3D Gaussians the renderings of which can be encoded into tokens to seamlessly connect with LLM.}
\label{fig:introfig}
\vspace{4mm}
\end{center}%
}]

\thispagestyle{empty}
% \vspace{-10mm}
% \begin{figure}[t]
% 	\centering
% 	\includegraphics[width=\linewidth]{fig/teaser.pdf}
%      \caption{\label{fig:teaser}ChatSplat is the first 3D Gaussian Splatting-based approach that enables conversational interaction with a 3D environment across multiple levels, including view, object, and the entire scene. The key idea is to learn a 3D conversational fields with 3D Gaussians who renderings can be encoded into tokens to seamlessly connect with LLM.
%      }
% \end{figure}
\maketitle

\begin{abstract} 
    Humans naturally interact with their 3D surroundings using language, and modeling 3D language fields for scene understanding and interaction has gained growing interest. This paper introduces ChatSplat, a system that constructs a 3D language field, enabling rich chat-based interaction within 3D space. Unlike existing methods that primarily use CLIP-derived language features focused solely on segmentation, ChatSplat facilitates interaction on three levels: objects, views, and the entire 3D scene. For view-level interaction, we designed an encoder that encodes the rendered feature map of each view into tokens, which are then processed by a large language model (LLM) for conversation. At the scene level, ChatSplat combines multi-view tokens, enabling interactions that consider the entire scene. For object-level interaction, ChatSplat uses a patch-wise language embedding, unlike LangSplat’s pixel-wise language embedding which implicitly includes mask and embedding. Here, we explicitly decouple the language embedding into separate mask and feature map representations, allowing more flexible object-level interaction. To address the challenge of learning 3D Gaussians posed by the complex and diverse distribution of language embeddings used in the LLM, we introduce a learnable normalization technique to standardize these embeddings, facilitating effective learning. Extensive experimental results demonstrate that ChatSplat supports multi-level interactions—object, view, and scene—within 3D space, enhancing both understanding and engagement.
    
    % significantly outperforms the previous state-of-the-art method LERF by a large margin. Notably, LangSplat is extremely efficient, achieving a 199 × speedup compared to LERF at the resolution of 1440 × 1080. Our code will be released after acceptance.
    
\end{abstract} \vspace{-4mm}
\section{Introduction}
\label{sec:intro}
    Natural language is the primary medium through which humans communicate~\cite{bonvillain2019language}. Modeling a 3D language field enables users to engage in an interactive dialogue with 3D environments, providing a powerful approach to enhance human-computer interaction and environmental understanding~\cite{azuma2022scanqa,cascante2022simvqa,gordon2018iqa}. Interest in modeling language fields in 3D has grown significantly due to its wide range of applications, including robotic navigation~\cite{huang2023visual} and manipulation~\cite{shen2023distilled}, 3D semantic comprehension~\cite{chen2023open,zhi2021place} and editing~\cite{kobayashi2022decomposing}, autonomous driving~\cite{jatavallabhula2023conceptfusion}, as well as augmented/virtual reality~\cite{liu2023weakly}.
    
    The lack of large-scale, diverse 3D scene datasets with language annotations has driven current methods (\eg, LERF~\cite{kerr2023lerf}) to rely on feature distillation from vision-language models %like
    such as CLIP to embed this information into 3D scenes. Building upon this, LangSplat~\cite{qin2024langsplat} integrates CLIP text embeddings into 3D Gaussian Splatting~\cite{kerbl20233dgs} to achieve highly efficient processing. However, these methods focus primarily on open-ended language queries in 3D, mainly supporting open-vocabulary segmentation. While these approaches are termed ``language fields'', their capabilities are limited to object query instead of interactive ``chatting'' with objects for comprehensive 3D scene understanding.

    In this work, we introduce 3D Conversational Gaussian Splatting (ChatSplat), a novel approach that embeds language within Gaussian Splatting by optimizing embeddings from vision-language models %like
    such as LLaVA~\cite{liu2023llava,li2024llava} directly into 3D scenes. ChatSplat extends beyond open-vocabulary segmentation, enabling conversational interaction with objects, views, and entire scenes in 3D to support a deeper understanding of the 3D environment. Notably, our ChatSplat utilizes LLaVA without the need for fine-tuning, thereby facilitating interaction at multiple levels: engaging with views, 3D scenes, and individual objects within a 3D space. To achieve view-level interaction, we design an encoder that converts %each view’s 
    the rendered feature map of each view into tokens, which are then processed by a large language model (LLM) for conversational responses. At the scene level, our ChatSplat combines tokens from multiple views, encoded by our encoder %\fw{\hl{fangyin: an encoder that decodes?}} %, 
    to enable interactions that consider the entire scene. For object-level interaction, \gh{our} ChatSplat employs a patch-wise language embedding %, 
    in contrast to %LangSplat’s 
    \gh{the} pixel-wise embedding \gh{in LangSplat} that merges mask and embedding information. Here, we explicitly separate the language embedding into distinct mask and feature map representations %, enabling
    \gh{to enable} more versatile object-level interactions. Specifically, we render a mask map and a view feature map %, 
    \gh{and} then isolate an object feature map based on the relevant object mask. We then decode the object feature map into tokens, distilling language knowledge into these tokens. %Besides
    \gh{Furthermore}, the distribution of language embeddings used in the LLM is usually complex and \fw{heterogeneous}, which makes it difficult for Gaussian Splatting to learn directly. To address the challenge, we introduce an autoencoder to first normalize these embeddings, facilitating effective learning. 
    We summarize %the 
    \gh{our main} contributions of this paper as follows:
    \begin{itemize}
    \item We introduce ChatSplat, the first 3D Gaussian Splatting-based approach enabling conversational interaction with a 3D environment \fw{across multiple levels}.
    
    \item We develop an encoder that \fw{encodes} rendered feature maps into language tokens and explicitly decouples object language embeddings into separate object masks and feature maps. Additionally, we introduce a scene-specific autoencoder to alleviate the complex and varied distribution of language embeddings.

    \item Experimental results demonstrate that our method enables effective conversational interaction with objects, views, and scenes in a 3D environment.
    \end{itemize}

\begin{figure*}[t]
	\centering
	\includegraphics[width=\linewidth]{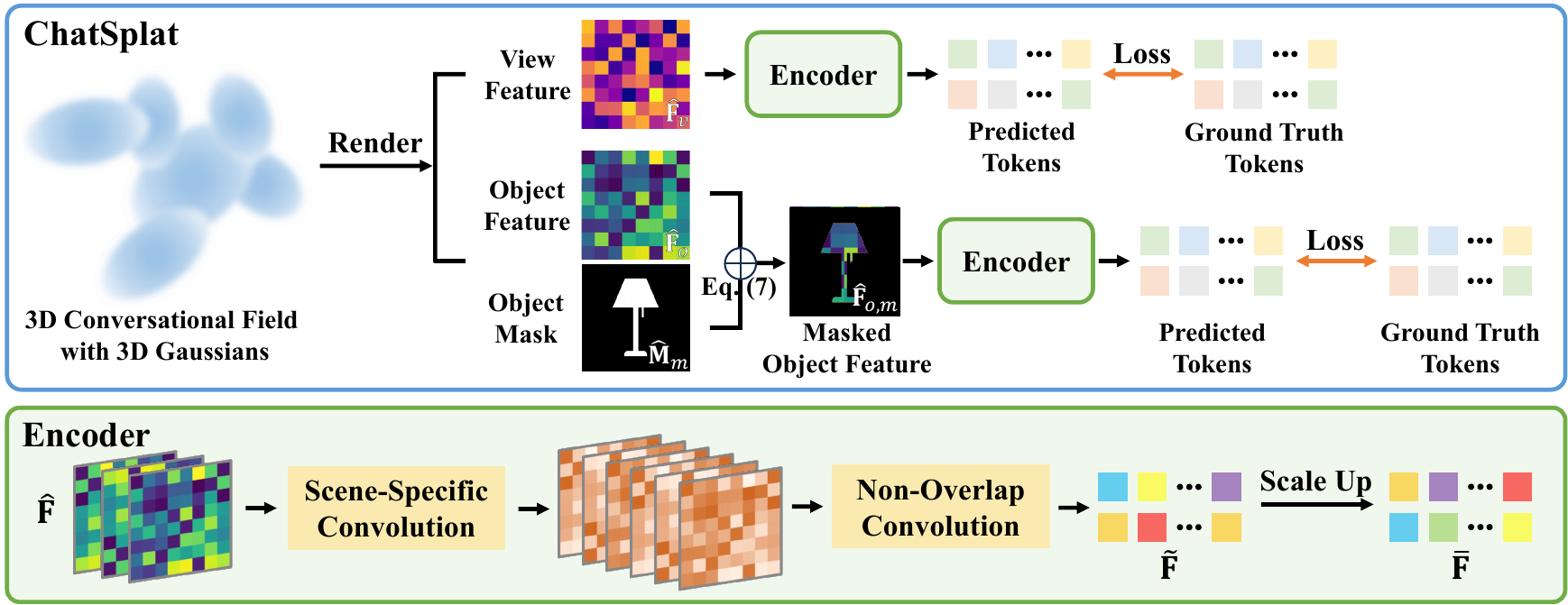}
     \caption{\label{fig:framework}\textbf{Overview of ChatSplat.} Our framework generates hierarchical language feature maps to support view-, object-, and scene-level chatting. For object-level chatting, we first render a mask to isolate the feature map of the selected object. \fw{The proposed encoder then converts the feature map into LLM's input through dimensionality lifting and tokenization}.
     }
     % \vspace{-0.4cm}
\end{figure*}

\section{Related Work}
\label{sec:related}
\subsection{3D Gaussian Splatting}
    Real-time rendering has long been a crucial objective in neural rendering research. Recently, Kerbl et al.~\cite{kerbl20233dgs} introduced \fw{3D Gaussian Splatting (3DGS)} to represent a 3D scene using a collection of 3D Gaussians, achieving real-time rendering at high resolution \fw{with} state-of-the-art visual quality. \fw{The success of 3DGS} has inspired numerous extensions to other tasks. \fw{Several studies~\cite{wu20234d,luiten2023dynamic,yang2023real,duan:2024:4drotorgs,duan20244d} have adapted 3D Gaussian Splatting for dynamic scenes}. For 3D surface reconstruction, some works~\cite{chen2023neusg,chen2024vcr} flatten 3D \fw{Gaussians into 2D planes to align them} with the surface. \chl{Recently, other studies~\cite{ye2024gs-grouping,qin2024langsplat} have utilized 3DGS for segmentation. Further discussions on this topic can be found in \secref{sec:language_field}.} However, these methods primarily focus on segmentation, despite integrating language \fw{that carries much richer} information. In contrast, our work expands each 3D Gaussian with language embeddings to enhance high-level scene understanding, enabling not only segmentation but also interactive ``chatting'' with the scene.

% For example, Gaussian Group~\cite{ye2024gs-grouping} leverages semantic Gaussians using open-world segmentation masks generated by DEVA~\cite{cheng2023tracking}. Additionally, LangSplat~\cite{qin2024langsplat} incorporates \fw{CLIP~\cite{radford2021clip} text embeddings} into Gaussians, enabling open-vocabulary segmentation through a learned language field.\fw{\hl{Fangyin: These few papers like Gaussian Grouping, langSplat are also discussed in Sec. 2.3 3D Language Field, shall we move and merge all discussions into Sec. 2.3 to reduce redundancy? We can summarize these into one sentence and point readers to Sec. 2.3 instead.}} 
    
\subsection{3D Semantic Field}
    Semantic-NeRF~\cite{semantic_nerf} pioneered the integration of semantics into Neural Rendering Fields~\cite{deng2022ds-nerf}, laying the groundwork for further research in this area. Building on this concept, various studies have incorporated additional features. For example, some works~\cite{fu2022panoptic, kundu2022panoptic, kobayashi2022distilledfeaturefields} integrate instance modeling, while others~\cite{tschernezki22neural} encode abstract visual features, allowing for post hoc semantic segmentation. Panoptic NeRF~\cite{fu2022panoptic} and DM-NeRF~\cite{wang2022dmnerf} focus on panoptic radiance fields, designed for applications like label transfer and scene editing. GNeSF~\cite{chen2023gnesf} introduces a generalizable semantic field, targeting the challenge of generalization to unseen scenes. More recently, GOV-NeSF~\cite{wang2024gov} proposes a generalizable implicit representation for 3D scenes with open-vocabulary semantics, further advancing the field of semantic 3D scene representation.
    
\subsection{3D Language Field}
    \label{sec:language_field}
    Beyond the semantic field, LERF~\cite{kerr2023lerf} was the first to incorporate CLIP~\cite{radford2021clip} features into NeRF, enabling open-vocabulary 3D queries through the powerful CLIP~\cite{radford2021clip} representation. OpenNeRF~\cite{engelmann2024opennerf} introduces a novel approach for generating diverse camera poses, enhancing feature extraction. Among 3DGS techniques, LangSplat~\cite{qin2024langsplat} uses a two-stage strategy for semantic feature acquisition, while Gaussian Grouping~\cite{ye2024gs-grouping} adapts Gaussian splatting for combined reconstruction and segmentation. Additionally, GOI~\cite{qu2024goi} uses hyperplane division to improve feature alignment, and 3DVLGS~\cite{peng20243d} proposes a cross-modal rasterizer that fuses modalities and employs a smoothed semantic indicator to enhance semantic rasterization. Although these methods incorporate language knowledge, they predominantly focus on segmentation tasks. In contrast, our work extends each 3D Gaussian with language embeddings to support \fw{comprehensive} high-level scene understanding, facilitating not only segmentation but also interactive ``chatting'' with the scene.

\section{Our Method}
\label{sec:method}

In this section, we first revisit the challenges of modeling 3D conversational fields and then elaborate on how our proposed ChatSplat addresses these issues. \figref{fig:framework} depicts the framework of our proposed ChatSplat.

\subsection{The Challenges of Conversational Fields}
    \label{sec:challenge}

    % In contrast to LangSplat’s pixel-wise features, ChatSplat applies patch-level features. In detail, pixel-wise embedding merges mask and embedding information. Here, we explicitly separate the language embedding into distinct mask and feature map representations, enabling more versatile object-level interactions. Specifically, we render a mask map and a view feature map, then isolate an object feature map based on the relevant object mask.

    % besides the rendered object-level language feature map, 
    
    Most existing methods~\cite{kerr2023lerf,liu2023weakly,shafiullah2022clip} \fw{extract image feature from CLIP~\cite{radford2021clip}} to supervise the 3D language field. These methods leverage the well-aligned text-image latent space provided by CLIP~\cite{radford2021learning,li2022ordinalclip,zhou2022learning,qin2024langsplat} %, which facilitates 
    \fw{for} open-vocabulary queries. %In particular
    \gh{For example}, LangSplat~\cite{qin2024langsplat} combines SAM~\cite{kirillov2023sam} and CLIP~\cite{radford2021clip} to perform open-vocabulary segmentation queries. However, their functionality is limited to segmentation tasks. In contrast, our ChatSplat extends this capability
    %, moving 
    beyond segmentation to enable conversational interactions or ``chat'' within 3D scenes.

    \gh{The shift in focus from open-vocabulary segmentation queries to the development of} %Developing 
    conversational fields for 3D environments %presents
    \gh{possesses two} %several 
    key challenges\fw{:}%from
    \gh{ the structural differences between feature representations and the additional complexity of the language embeddings in LLMs.} %In particular, the structural differences between feature representations in LangSplat and our ChatSplat, as well as the complexity of language embeddings in large language models (LLMs), necessitate specific solutions.
    
    Firstly, the structure of language embeddings \gh{our} \fw{ChatSplat differ significantly from that of} \gh{open-vocabulary segmentation queries works such as} LangSplat~\cite{qin2024langsplat}. In LangSplat~\cite{qin2024langsplat}, the language embedding is pixel-wise, meaning that %each pixel’s feature 
    \gh{the feature of each pixel} independently represents its associated object. This setup allows for straightforward object representation as %each pixel’s feature \gh{the features of each pixel} is independent of the features of other pixels. 
    \fw{the features of different pixels are independent of each other}. In contrast, ChatSplat uses patch-wise embeddings, where features are organized as feature maps that represent entire views or objects. This structure simplifies view representation but complicates object representation. Unlike LangSplat~\cite{qin2024langsplat} where features implicitly represent objects, ChatSplat %requires
    \fw{explicitly decouples} language embeddings into separate mask and feature map representations for each object. This explicit representation ensures accurate and meaningful language embeddings for each object in the 3D scene.
    
    Secondly, the distribution of language embeddings in LLMs is considerably more complex and \fw{heterogeneous} than in models like CLIP~\cite{radford2021clip} due to the broader range of tasks they perform. 
    % CLIP’s language embeddings, designed primarily for classification tasks, are limited to a range of approximately -1 to +1.
    \gh{The language embeddings of CLIP are designed primarily for classification tasks, and thus are limited to a range of approximately $-1$ to $+1$.}
    In contrast, the embeddings in LLMs %which are used 
    \fw{trained} for more intricate conversational tasks span a range from roughly $-100$ to $+100$, reflecting the nuanced linguistic context that LLMs capture. This wider range is essential for supporting rich dialogues but presents challenges new for embedding language features within \gh{spatial representations such as} 3D Gaussians. To address this \gh{challenge}, we introduce a scaling adjustment to align the language features of Gaussians with the expanded embedding range of LLMs %, ensuring 
    \gh{to ensure} that our ChatSplat can effectively support language-based interactions within the 3D environment.

\subsection{Conversational Gaussian Splatting}
    \label{sec:splat}

    \fw{3DGS}~\cite{kerbl20233dgs} \fw{explicitly} represents a 3D scene structure using 3D Gaussians. Each Gaussian is defined by a covariance matrix $\boldsymbol{\Sigma}$ and a center point $\mathbf{p} \in \mathbb{R}^3$
    %, representing the Gaussian's mean
    \gh{that represents the Gaussian mean}. The distribution of the 3D Gaussian is given by:
    \begin{equation}
        \label{eq:gaussian_dist}
        G(\mathbf{x}) = \exp{\left\{-\frac{1}{2}(\mathbf{x}-\mathbf{p})^\top \boldsymbol{\Sigma}^{-1}(\mathbf{x}-\mathbf{p})\right\}}.
    \end{equation}
    To ensure positive semi-definiteness during optimization, the covariance matrix $\boldsymbol{\Sigma}$ is expressed as a combination of a scaling matrix $\mathbf{S}$ and a rotation matrix $\mathbf{R}$:
    \begin{equation}
        \boldsymbol{\Sigma} = \mathbf{R} \mathbf{S} \mathbf{S}^\top \mathbf{R}^\top,
    \end{equation}
    where $\mathbf{S}$ is a diagonal matrix represented by scaling factors $\mathbf{s} \in \mathbb{R}^3$, and $\mathbf{R}$ is parameterized as a quaternion $\mathbf{r} \in \mathbb{R}^4$.
    
    For novel view rendering, splatting~\cite{yifan2019differentiable} is applied to project Gaussians onto camera planes. Using the viewing transform matrix $\mathbf{W}$ and the Jacobian $\mathbf{J}$ of the projective transformation~\cite{zwicker2001surface}, the transformed covariance matrix $\boldsymbol{\Sigma}'$ is calculated as:
    \begin{equation}
        \boldsymbol{\Sigma}' = \mathbf{J} \mathbf{W} \boldsymbol{\Sigma} \mathbf{W}^\top \mathbf{J}^\top.
    \end{equation}
    A 3D Gaussian is thus defined by its position $\mathbf{p}$, quaternion $\mathbf{r}$, scaling factor $\mathbf{s}$, opacity $o \in \mathbb{R}$, and color given by spherical harmonics coefficients $\mathbf{H} \in \mathbb{R}^k$. For each pixel, color and opacity contributions from multiple Gaussians are blended using:
    \begin{equation}
        \hat{\mathbf{C}}(k) = \sum_{i \in N} \mathbf{c}_i \alpha_i \prod_{j=1}^{i-1} (1 - \alpha_j),
    \end{equation}
    where $\mathbf{c}_i$ and $\alpha_i = o_i G(\textbf{x}_i)$ are the color and density of a Gaussian, respectively, and $\hat{\mathbf{C}}(k)$ is the final rendered color for pixel $k$.
    
    % Beyond color rendering, we introduce 3D Conversational Gaussian Splatting, augmenting each 3D Gaussian with view-level and object-level language embeddings $\{\mathbf{f}_v, \mathbf{f}_o\}$ derived from LLaVA features, forming 3D language Gaussians. 
    \gh{To achieve our 3D Conversational Gaussian Splatting, we go beyond color rendering by augmenting each 3D Gaussian with additional view-level and object-level language embeddings $\{\mathbf{f}_v, \mathbf{f}_o\}$ derived from LLaVA~\cite{liu2023llava,li2024llava} features to form 3D language Gaussians.}
    For efficiency, we use a tile-based rasterizer, and render language features with:
    \begin{equation}
        \hat{\mathbf{F}_l}(k) = \sum_{i \in N} \mathbf{f}_{l,i} \alpha_i \prod_{j=1}^{i-1} (1 - \alpha_j), \quad l \in \{v, o\},
        \label{eq:render_feature}
    \end{equation}
    where $\hat{\mathbf{F}}_l(k)$ is the rendered language feature for pixel $k$ at level $l$ in the feature map $\hat{\mathbf{F}}_l \in \mathbb{R}^{D \times H \times W}$. Here, $D$ denotes the channel dimension of the feature map, while $H$ and $W$ correspond to its height and width, respectively. By embedding language directly into \gh{the} Gaussians, we enable 
    %the 3D language field to respond to language-based questions.
    \gh{the response of 3D language field to language-based questions.}
    
    For view-level interaction, we encode the view-level feature map $\hat{\mathbf{F}}_v$ into tokens using the encoder detailed in the following section, which are then processed by an LLM for conversational interaction. Object-level interactions are more complex, requiring extraction of object-specific language features from the object-level feature map $\hat{\mathbf{F}}_o$. %To do this
    \gh{To this end}, we render an object mask $\hat{\mathbf{M}} \in \mathbb{R}^{M \times H \times W}$ %, where $M$ represents the number of objects, 
    to mask out the relevant object-level feature map\gh{, where $M$ represents the number of objects}. The mask is computed as:
    \begin{equation}
        \hat{\mathbf{M}} = \sum_{i \in N} \mathbf{m}_i \alpha_i \prod_{j=1}^{i-1} (1 - \alpha_j),
    \end{equation}
    where $m_i$ denotes the identity of each Gaussian. Using this mask, the object-specific feature map $\hat{\mathbf{F}}_{o,m}$ is obtained as:
    \begin{equation}
        \hat{\mathbf{F}}_{o,m} = \text{MaskOut}(\hat{\mathbf{F}}_o , \hat{\mathbf{M}}_m),
        \label{eq:maskout}
    \end{equation}
    where $\hat{\mathbf{M}}_m \in \mathbb{R}^{H \times W}$ is the mask for object $m$. The \text{MaskOut} function isolates the language feature map for object $m$ from the object-level feature map, enabling targeted object-level interactions.

\subsection{Encoder}
    \label{sec:encoder}

    Our encoder $\operatorname{Encoder}(\cdot)$ primarily consists of two types of convolutional networks: scene-specific convolution networks for dimension lifting and a non-overlapping convolutional network for tokenization.

    As an explicit modeling approach, \gh{our} ChatSplat may generate millions of 3D points to represent a detailed 3D scene. Given the high dimensionality of language embeddings, directly learning $\mathbf{f}_l$ in the language latent space significantly increases memory and computational demands. While learning RGB colors without spherical harmonics has relatively low memory requirements, learning 1,152-dimensional language features increases the memory for storing 3D Gaussians by about 100-fold, often exhausting L1 cache memory.
    
    To \fw{reduce the memory and improve efficiency}, %address this\gh{the issue of high memory requirement}, 
    we introduce a scene-specific convolutional network that maps high-dimensional language embeddings into a more compact latent space. % \gh{for memory usage reduction and efficiency enhancement.}%, reducing memory usage and enhancing efficiency. 
    The %LLM’s 
    $D$-dimensional latent space \gh{of LLM} is inherently compact as it is designed to process extensive text data. However, our language field is scene-specific, allowing us to leverage scene priors to compress language features effectively. Each input image yields hundreds of object masks, which represent far fewer samples than the training data used for LLMs. This sparse distribution of segmented regions within the LLM latent space enables further compression of language features through our scene-specific convolutional network.
    
    After generating the high-dimensional feature map, we convert it into tokens to match the input structure of the LLM. Specifically, we employ a non-overlapping convolutional network to transform the feature map into tokens, ensuring effective alignment with the LLM processing framework. This process can be represented as:
    \begin{equation}
        \tilde{\textbf{F}} = \operatorname{Encoder}(\hat{\textbf{F}}),
    \end{equation}
    where $\tilde{\textbf{F}}$ is the encoded language feature map, which is either the view-level feature map $\tilde{\mathbf{F}}_{v}$ %\fw{\hl{Fangyin: is it tilde or hat?}} 
    or the object-level feature map $\tilde{\mathbf{F}}_{o,m}$.%\fw{\hl{Fangyin: is it tilde or hat?}}.
    
    \paragraph{Scaling adjustment.} 
    %However
    \gh{Unfortunately}, integrating these tokens into the LLM %presents a challenge 
    \gh{is non-trivial} due to the complex and diverse distribution of language embeddings within the LLM latent space. 
    To address this challenge, we apply a scaling and shifting operation to align the tokens with the target token distribution:
    \begin{equation}
        \bar{\textbf{F}} = \tilde{\textbf{F}} \cdot \textbf{a} + \textbf{b},
    \end{equation}
    where $\textbf{a} \in \mathbb{R}^{T \times D}$ is a learnable scaling factor, and $\textbf{b} \in \mathbb{R}^{T \times D}$ is a learnable shift. This adjustment ensures that the encoded tokens are compatible with the scale of the LLM tokens, enhancing the integration and facilitating effective language-based interaction.

% \subsection{Decouple}
\subsection{Loss Function}

    To train our 3D language field effectively, we define two primary loss functions: a view-level loss $\mathcal{L}_{\text{v}}$ and an object-level loss $\mathcal{L}_{\text{o}}$. These losses are designed to ensure that the generated language features align closely with the target features at both the view and object levels.

    The view-level loss $\mathcal{L}_{\text{v}}$ is formulated as the L1 norm between the predicted view-level feature map $\bar{\textbf{F}}_v$ and the target feature map $\textbf{F}$:
    \begin{equation}
        \mathcal{L}_{\text{v}} = \|\bar{\textbf{F}}_v - \textbf{F}\|_1.
        \label{eq:loss_v}
    \end{equation}
    This loss encourages the model to capture accurate language features at the view level, helping to ensure that the scene representation is consistent with the desired feature distribution.
    
    For object-level features, we define $\mathcal{L}_{\text{o}}$ as the mean L1 norm across all object-specific feature maps $\bar{\textbf{F}}_{o,m}$ compared to their respective target feature maps $\textbf{F}_m$, where $M$ is the total number of objects:
    \begin{equation}
        \mathcal{L}_{\text{o}} = \frac{1}{M} \sum_{m \in M} \|\bar{\textbf{F}}_{o,m} - \textbf{F}_m\|_1.
        \label{eq:loss_o}
    \end{equation}
    This object-level loss is essential for object-specific interactions, allowing %us 
    \gh{a user} to ``chat'' with individual objects within the 3D scene.
    
    The total loss $\mathcal{L}_{\text{total}}$ is defined as the sum of the view-level and object-level losses:
    \begin{equation}
        \mathcal{L}_{\text{total}} = \mathcal{L}_{\text{v}} + \mathcal{L}_{\text{o}}.
    \end{equation}
    This combined loss function guides the network to learn accurate language features across both the view and object levels, supporting effective language-based interactions throughout the 3D environment.

\subsection{Chatting in Scenes}

    To enable conversational interaction within the 3D environment, we follow a series of steps to render and process language embeddings from 3D to 2D, allowing both view-level and object-level chats:

    \begin{enumerate}
        \item During chatting, we render the language embeddings from 3D to 2D according to Eq.~\eqref{eq:render_feature}. This projection creates 2D feature maps containing the relevant language information for each view or object within the scene.
        \item For \textbf{\textit{view-level chat}}, we utilize the trained decoder to convert the feature map of a view into tokens that represent the %view’s 
        language context \gh{of the view}. For \textbf{\textit{scene-level chat}}, multi-view feature maps are fed into the decoder to generate tokens that collectively represent the 
        %entire scene’s 
        language context \gh{of the entire scene}. For \textbf{\textit{object-level chat}}, users can first select an object by clicking on it within the 3D scene. Once the object is selected, we apply Eq.~\eqref{eq:maskout} to extract the object-specific feature map. This feature map is then processed by the encoder to produce tokens that capture the language features of the selected object, enabling conversational interaction at the object level.
        \item Users can input questions on the view or selected object, which are fed into an LLM along with the corresponding tokens. This setup enables interactive dialogue with the view or individual objects, facilitating meaningful language-based interactions within the 3D environment.
    \end{enumerate}
    
    % First, during chatting, we render the language embeddings from 3D to 2D according to Eq.~\eqref{eq:render_feature}. This projection creates 2D feature maps containing the relevant language information for each view or object within the scene.
    
    % For view-level chatting, we utilize the trained decoder to convert the feature map of a view into tokens that represent the %view’s 
    % language context \gh{of the view}. For scene-level chatting, multi-view feature maps are fed into the decoder to generate tokens that collectively represent the 
    % %entire scene’s 
    % language context \gh{of the entire scene}. For object-level chatting, users can first select an object by clicking on it within the 3D scene. Once the object is selected, we apply Eq.~\eqref{eq:maskout} to extract the object-specific feature map. This feature map is then processed by the encoder to produce tokens that capture the language features of the selected object, enabling conversational interaction at the object level.

    % Finally, users can input questions regarding the view or selected object, which are fed into an LLM along with the corresponding tokens. This setup enables interactive dialogue with the view or individual objects, facilitating meaningful language-based interactions within the 3D environment.

\section{Experiments}
\label{sec:exp}

    \begin{figure*}[h]
    	\centering
    	\includegraphics[width=\linewidth]{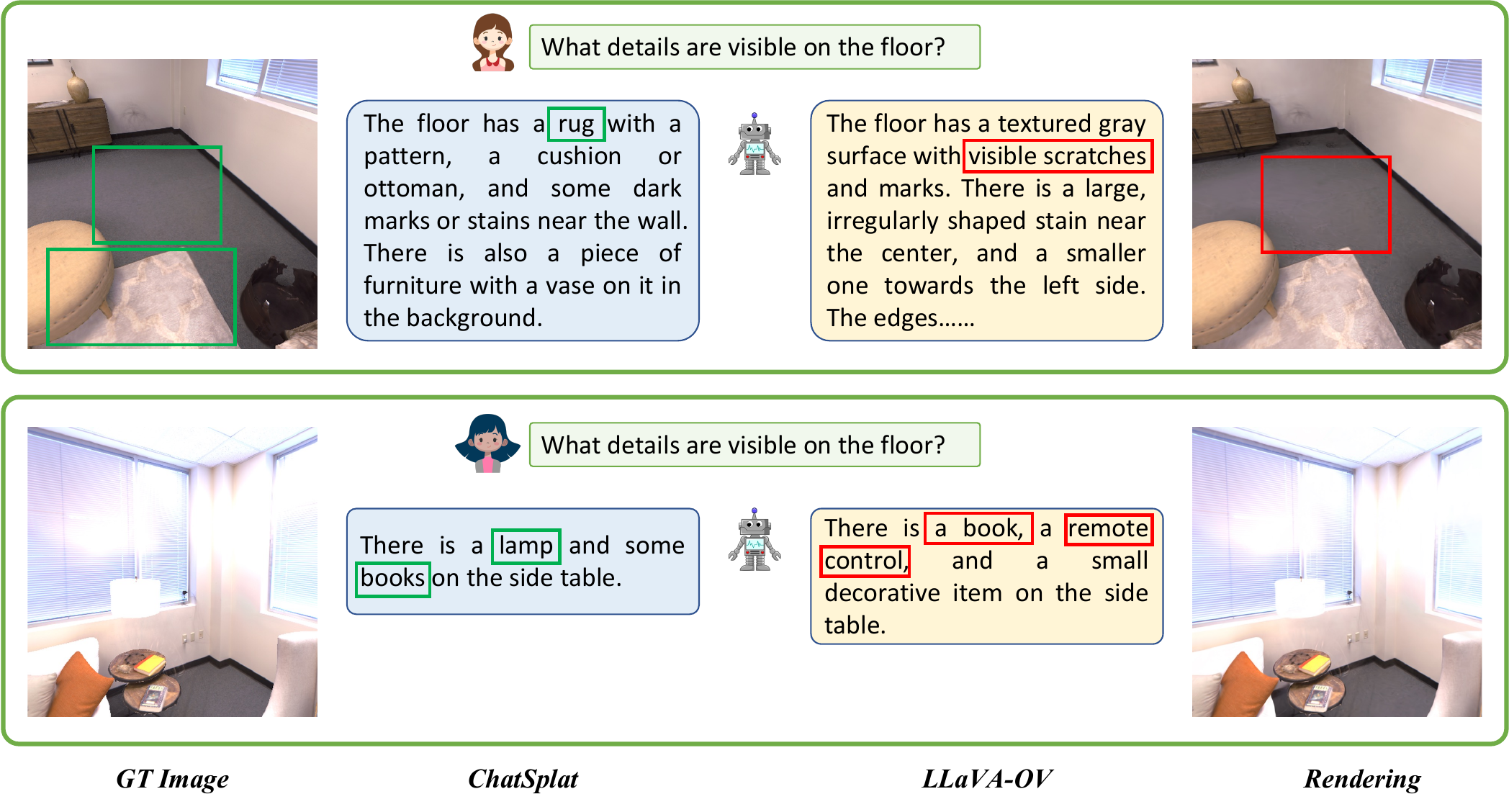}
        \vspace{-0.7cm}
         \caption{\textbf{Qualitative comparison on view-level chatting.} \fcolorbox{green}{white}{Correct answers} are highlighted with green boxes, while \fcolorbox{red}{white}{incorrect ones} are marked in red. Our ChatSplat outperforms the baseline method, LLaVA-OV, primarily because LLaVA-OV relies solely on rendered images. These rendered images inherently contain errors, which are propagated to LLaVA-OV, leading to compounded inaccuracies in the final output. Specifically, as shown in the first row of \figref{fig:view}, LLaVA-OV incorrectly outputs "visible scratches" due to rendering errors, with scratches erroneously appearing in the rendered image on the right.
         }
         \label{fig:view}
    \end{figure*}

    % \begin{table*}[ht]
%     \centering
%     \renewcommand{\arraystretch}{1.5} % 设置行间距为原来的1.5倍
%     \begin{tabular}{cccccccccc} % 增加一列
%     \toprule
%     \multirow{2}{*}{\textbf{Method}} & \multicolumn{2}{c}{\textbf{LERF}} & \multicolumn{5}{c}{\textbf{Replica}} & \multirow{2}{*}{\textbf{Mean}} & \multirow{2}{*}{\textbf{Speed} (FPS)} \\
%     \cmidrule(lr){2-3} \cmidrule(lr){4-8}
%      & \textbf{teatime} & \textbf{figurines} & \textbf{room\_0} & \textbf{room\_1} & \textbf{office\_0} & \textbf{office\_2} & \textbf{office\_3} & & \\
%     \midrule
%     LLaVA-OV & & & & & & & & & 11 \\
%     ChatSplat & & & & & & & & & 73 \\
%     \bottomrule
%     \end{tabular}
%     \caption{\textbf{Comparison of LLaVA-OV and ChatSplat across LERF and Replica datasets on view-level chatting.}}
%     \label{tab:view}
% \end{table*}

\begin{table*}[h!]
    \centering
    \begin{tabular}{cccccccc}
    \toprule
    \multirow{2}{*}{\textbf{Method}} & \multicolumn{3}{c}{\textbf{View}} & \multicolumn{3}{c}{\textbf{Scene}} & \multirow{2}{*}{\textbf{FPS}} \\
    \cmidrule(lr){2-4} \cmidrule(lr){5-7}
     & \textbf{teatime} & \textbf{ramen} & \textbf{Mean} & \textbf{teatime} & \textbf{ramen} & \textbf{Mean} & \\
    \midrule
    LLaVA-OV & 86.58 & 78.06 & 82.32 & 85.00 & 90.00 & 87.50 & 9 \\
    ChatSplat & \textbf{87.37} & \textbf{84.19} & \textbf{85.78} & \textbf{86.25} & \textbf{91.67} & \textbf{88.96} & \textbf{72} \\
    \bottomrule
    \end{tabular}
    \caption{\textbf{Quantitative Results on the LERF Dataset at View and Scene Level.} 
    % We evaluate our method using GPT-4o scores and FPS as metrics. 
    \fw{We report the GPT-4o scores and FPS to evaluate the conversational quality and running efficiency, respectively. The baseline renders novel views as inputs to LLaVA-OV~\cite{li2024llava} for chatting. }
    Our approach %significantly outperforms These results highlight the superior , 
    achieves notable improvements in GPT-4o scores while maintaining significantly higher inference speed (real-time).}
    \label{tab:lerf_vs}
\end{table*}

    \begin{table*}[h!]
    \centering
    \begin{tabular}{cccccccccccc}
    \toprule
    \multirow{2}{*}{\textbf{Method}} & \multicolumn{5}{c}{\textbf{View}} & \multicolumn{5}{c}{\textbf{Scene}} & \multirow{2}{*}{\textbf{FPS}} \\
    \cmidrule(lr){2-6} \cmidrule(lr){7-11}
     & \textbf{room\_0} & \textbf{room\_1} & \textbf{office\_2} & \textbf{office\_3} & \textbf{Mean} & \textbf{room\_0} & \textbf{room\_1} & \textbf{office\_2} & \textbf{office\_3} & \textbf{Mean} & \\
    \midrule
    LLaVA-OV & 81.33 & \textbf{83.62} & 85.00 & 83.12 & 83.27 & 76.50 & 73.33 & 75.83 & \textbf{82.50} & 77.04 & 11 \\
    ChatSplat & \textbf{84.17} & 83.53 & \textbf{85.53} & \textbf{84.06} & \textbf{84.32} & \textbf{80.50} & \textbf{75.00} & \textbf{80.71} & 81.67 & \textbf{79.47} & \textbf{81} \\
    \bottomrule
    \end{tabular}
    \caption{\textbf{Quantitative Results on the Replica Dataset at View and Scene Level.} On the more complex indoor Replica dataset, our approach continues to outperform the baseline \fw{across most scenes and overall, in terms of conversational quality and inference speed.}
    %These results underscore the superior conversational quality and efficiency of our method, achieving substantial improvements in GPT-4o scores while sustaining extreme inference speed.
    }
    \label{tab:replica_vs}
     % \vspace{-0.1cm}
\end{table*}
    
We first introduce our experiment settings in \secref{sec:setting}. Then we evaluate Visual Question Answering in \secref{sec:vqa} from view, scene, and object levels. Additionally, we validate the effectiveness of the proposed techniques in \secref{sec:ablation}.

\subsection{Settings}
    \label{sec:setting}

    \textbf{Datasets.} We employ two datasets for evaluation: the LERF dataset~\cite{kerr2023lerf} and the Replica dataset~\cite{straub2019replica}. The LERF dataset, captured using the iPhone App Polycam, consists of complex in-the-wild scenes and was originally designed for 3D object localization and segmentation tasks. To extend its utility, we annotated ground truth for Visual Question Answering (VQA), enabling the evaluation of VQA as well as image and scene captioning tasks on this dataset. We report a GPT-4o~\cite{achiam2023gpt} score comparable to PointLLM~\cite{xu2024pointllm}. The Replica dataset is a photorealistic dataset curated for research in 3D vision and robotics. It features a collection of high-quality, richly annotated 3D indoor scenes with realistic spatial layouts, textures, and lighting conditions. Each scene provides dense semantic labels, accurate geometry, and high-resolution RGB-D data. Similar to the LERF dataset, we annotated ground truth for VQA, allowing the evaluation of VQA and image or scene captioning tasks

    \begin{figure*}[t]
    	\centering
    	\includegraphics[width=\linewidth]{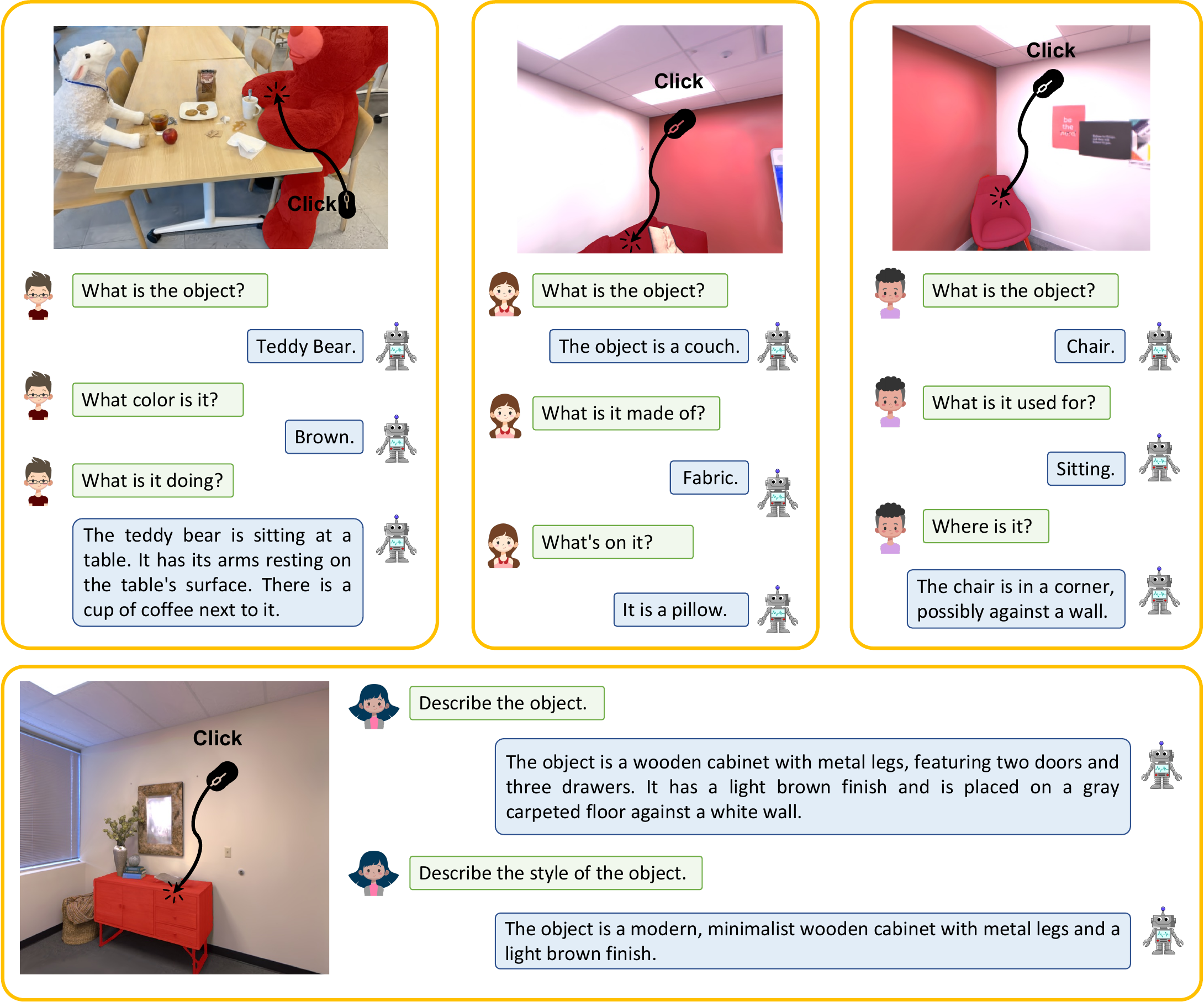}
         % \vspace{-0.3cm}
         \caption{\textbf{Qualitative results on object-level chatting.} For object-level chatting, the object is selected interactively using a mouse and highlighted in red for visual identification.
         }
         \label{fig:object}
         % \vspace{-0.4cm}
    \end{figure*}

    \vspace{4mm}
    \noindent\textbf{Implementation Details.}
    To extract language features for each image, we utilize the LLaVA-OV~\cite{li2024llava} 7B model. For each scene, we first train 3DGS~\cite{kerbl20233dgs} with RGB images and object labels, following the default parameter settings. For datasets with ground truth (GT) object labels, such as Replica, we use the provided labels directly. For datasets without GT object labels, such as LERF, we generate pseudo labels using the approach outlined in Gaussian-Group~\cite{ye2024gs-grouping}, leveraging DEVA~\cite{cheng2023tracking}. %The training process for 3DGS is conducted for 
    \fw{We train 3DGS} for 30,000 iterations to achieve robust representation. Subsequently, we train our 3D conversational Gaussians by fixing all other parameters of the 3D Gaussians, including position and opacity. During this stage, only the view-level and object-level language features are set as learnable, enabling effective adaptation for conversational tasks within the 3D environment. We train and test our method and baseline on NVIDIA A6000.

\subsection{Comparsion}
    \label{sec:vqa}
    As previously mentioned, we use GPT-4o as a grading assistant to evaluate the performance of our method. The score ranges from 0 to 100, where 0 represents a completely irrelevant or incorrect answer, and 100 represents a perfect response. For comparison, we evaluate our method against a naive alternative approach: first rendering novel views \fw{as inputs to} state-of-the-art LLaVA-OV~\cite{li2024llava} to answer questions %based on these rendered views. 
    This comparison highlights the advantages of our approach of integrating semantics into 3D feature fields over the VLM baseline.

    ChatSplat consistently outperforms LLaVA-OV across all metrics on the LERF dataset, as shown in \tabref{tab:lerf_vs}. For view-level tasks, ChatSplat achieves a mean GPT-4o score of 85.78, surpassing LLaVA-OV’s 82.32. Similarly, for scene-level tasks, ChatSplat demonstrates superior performance with a mean score of 88.96, compared to 87.50 for LLaVA-OV. These results highlight ChatSplat’s enhanced capability for both view-level and scene-level conversational understanding. 

    The superior performance of ChatSplat can be attributed to its ability to learn knowledge from multiple views, which enables a \fw{a comprehensive 3D understanding to produce}
    %voting mechanism for producing 
    correct answers. In contrast, the baseline with LLaVA-OV can only access rendered images,
    %naive alternative directly applies LLaVA-OV on rendered images, 
    where inherent rendering errors are passed to LLaVA-OV, compounding losses in the final output. \chl{The qualitative comparison is shown in \figref{fig:view}}. \fw{In the first example, the \textit{visible scratches} detected by LLaVA-OV are actually artifacts of the 3DGS rendering and do not appear in the GT image.} \chl{In the second example, ChatSplat correctly identifies the \textit{lamp} and \textit{books}, while the baseline fails and incorrectly detects a nonexistent \textit{remote control}.} Furthermore, ChatSplat is significantly more efficient, achieving 72 FPS thanks to its fast feature rendering. In contrast, the alternative achieves only 9 FPS as it requires extracting language features from rendered images using computationally expensive deep neural networks. These results demonstrate that ChatSplat is highly suitable for real-time applications.

    On the more complex indoor Replica dataset, ChatSplat maintains its superior performance, as shown in \tabref{tab:replica_vs}. It surpasses LLaVA-OV not only in GPT-4o scores (84.32 \vs 83.27 for view-level chatting and 79.47 \vs 77.04 for scene-level chatting) but also in inference speed (81 \vs 11 FPS). These results further demonstrate ChatSplat’s ability to handle intricate indoor scenes with high conversational quality and efficiency %, solidifying its advantages 
    for multi-level 3D conversational tasks.

    For object-level chatting, the baseline involves using SAM~\cite{kirillov2023sam} to first mask out an object, followed by using LLaVA to chat with the object. This approach is inherently inefficient, as it requires the integration of two foundation models, both of which rely on extensive computation for object masking and language feature extraction. In contrast, ChatSplat simplifies the process while achieving superior performance in terms of both GPT-4o scores and efficiency. Detailed quantitative results are provided in the supplementary material. \chl{As shown in \figref{fig:object}, ChatSplat consistently provides accurate answers across diverse object properties including name, color, material, purpose, and location. This demonstrates ChatSplat's ability to deliver precise and context-aware responses.}

\subsection{Ablation}
    \label{sec:ablation}
    
    We verify the effectiveness of various design choices on converstaional quality, including the Encoder and the Scaling Adjustment, using the Replica dataset~\cite{straub2019replica}, and report the corresponding GPT-4o scores. 
    
    \begin{table}[h]
  \renewcommand\tabcolsep{8pt}
  \centering
  \begin{tabular}{cc|c}
    \toprule
    \multicolumn{2}{c|}{\textbf{Module}} & \multirow{2}{*}{\textbf{Score}} \\
    \cmidrule{1-2}
    Encoder & Scaling Alignment &  \\
    \midrule
       & \Checkmark & OOM  \\
    \Checkmark & & 72.54 \\
    \Checkmark & \Checkmark & \textbf{85.78} \\
    \bottomrule
  \end{tabular}
  \caption{\textbf{Ablations on the LERF dataset on view-level chat.}}
  \label{tab:ablation}
  % \vspace{-0.1cm}
\end{table}

    First, we examine the impact of our encoder. Since non-overlapping convolution is essential for converting feature maps into tokens to align with the structure of the language embeddings in LLMs, we only evaluate the effectiveness of the scene-specific convolution in the encoder. As shown in the first row of \tabref{tab:ablation}, omitting the scene-specific convolution results in training failure due to out-of-memory issues. 
    
    Furthermore, the importance of the scaling adjustment is evident. Without the it, the score drops by 13.24 (\tabref{tab:ablation}, second row). This demonstrates that the scaling adjustment effectively addresses the complex distribution of language embeddings in LLMs, improving overall performance.

\section{Conclusion}
    \label{sec:conclusion}
    We presented ChatSplat, a system that constructs a 3D Conversational field to enable chat-based interactions within 3D spaces. Unlike prior methods focused on segmentation, ChatSplat facilitates multi-level interactions at the object, view, and scene levels. By introducing a patch-wise embedding approach and a learnable normalization technique, ChatSplat effectively handles the complexity of LLM-based language embeddings. Experimental results demonstrate its ability to enhance understanding and engagement within 3D environments, setting the stage for advanced conversational interactions in 3D spaces.

    \vspace{4mm}
    \noindent \textbf{Limitations.} 
    Like most Gaussian Splatting methods, ChatSplat requires high-quality multi-view captures with known camera parameters, which are not always readily available or easy to obtain. Additionally, for object-level chatting in scenes without ground truth object labels, ChatSplat, similar to other Gaussian Splatting methods like Gaussian Group for segmentation, relies on DEVA to generate high-quality pseudo labels. Poor-quality labels can lead to failures. Moreover, ChatSplat depends on LLaVA for generating language features; if LLaVA fails in any view of the object, ChatSplat’s performance will also be compromised.

\clearpage
{
    \small

    \bibliographystyle{ieeenat_fullname}
    % \bibliography{main}
}

\clearpage
\setcounter{page}{1}
% \maketitlesupplementary
    % \begin{figure*}[!th]
    % 	\centering
    % 	\includegraphics[width=\linewidth]{fig/scene.pdf}
    %      % \vspace{-0.3cm}
    %      \caption{\textbf{Qualitative results on scene-level chatting.} For scene-level chatting, the object is selected interactively using a mouse and highlighted in red for visual identification.
    %      }
    %      \label{fig:scene}
    %      % \vspace{-0.4cm}
    % \end{figure*}

    \twocolumn[{%
\renewcommand\twocolumn[1][]{#1}%
% \maketitle
\maketitlesupplementary
\vspace{1.5cm}
\vspace{-3\baselineskip}
% \vspace{-3\baselineskip}
\begin{center}
\centering
\setlength{\tabcolsep}{0.5pt}
\captionsetup{type=figure}
{\footnotesize
\renewcommand{\arraystretch}{0.5} 

\begin{tikzpicture}
\node (img) {
\begin{tabular}{c c c c c c}
 \includegraphics[width=1\linewidth]{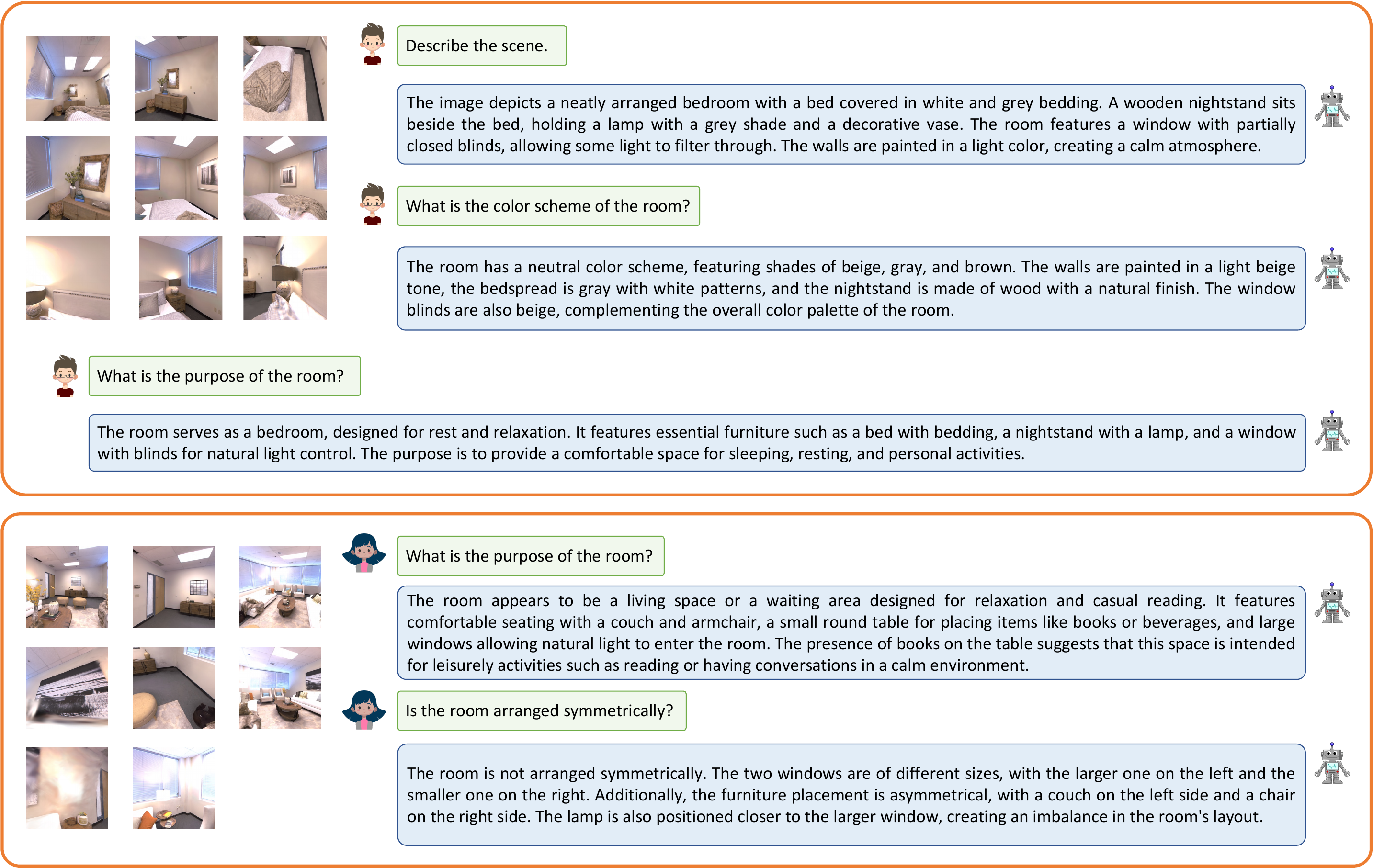}
\end{tabular}
};
\end{tikzpicture}

}
\vspace{-5mm}
\vspace{-0.5\baselineskip}
\hspace{20pt}\captionof{figure}{\label{fig:scene}\textbf{Qualitative results on scene-level chatting.} For scene-level chatting, multi-view feature maps are processed by the decoder to generate tokens that collectively represent the language context of the entire scene. These tokens are then input into an LLM to enable conversational interaction at the scene level.}
% \label{fig:introfig}
\vspace{4mm}
\end{center}%
}]

\thispagestyle{empty}
    % \begin{figure*}[!th]
    % 	\centering
    % 	\includegraphics[width=\linewidth]{fig/scene.pdf}
    %      % \vspace{-0.3cm}
    %      \caption{\textbf{Qualitative results on scene-level chatting.} For scene-level chatting, the object is selected interactively using a mouse and highlighted in red for visual identification.
    %      }
    %      \label{fig:scene}
    %      % \vspace{-0.4cm}
    % \end{figure*}
\appendix

    \section{Implementation Details}
        We conduct most experiments using PyTorch 2.1.2 and CUDA 12.4 on an NVIDIA A6000 GPU with 46 GB of memory. Most hyperparameters are set identically to those used in Gaussian Splatting~\cite{kerbl20233dgs}. The object-level and view-level language features of a 3D Gaussian are configured with 32 channels each, while the object identity feature is set to 16 channels. The learning rates for both the language features of 3D Gaussians and the encoder are set to 0.05. Additionally, the kernel size for the non-overlapping convolution is configured as 14, converting a $14 \times 14$ patch into a single token.

    \section{Conversational Dataset Preparation}
        As described in the main paper, we use the LERF~\cite{kerr2023lerf} and Replica~\cite{straub2019replica} datasets to evaluate our method. To adapt these datasets for the visual question answering (VQA) task, we extend them by annotating ground truth. Specifically, we first use GPT-4o~\cite{achiam2023gpt} to generate questions and corresponding answers about objects, views, or scenes. The prompt used to guide GPT-4o is: ``Design questions for the image and provide the answers. Questions must include an object caption question to describe the object, along with other related questions.'' We then manually review the generated questions and answers, removing any unreasonable or incorrect annotations to ensure the quality and accuracy of the ground truth data.
    
        \begin{table*}[bth]
    \centering
    \begin{tabular}{lccccccccc}
    \toprule
    \multirow{2}{*}{\textbf{Method}} & \multicolumn{5}{c}{\textbf{Replica}} & \multicolumn{3}{c}{\textbf{LERF}} & \multirow{2}{*}{\textbf{OBS}} \\
    \cmidrule(lr){2-6} \cmidrule(lr){7-9}
     & \textbf{room\_0} & \textbf{room\_1} & \textbf{office\_2} & \textbf{office\_3} & \textbf{Mean} & \textbf{teatime} & \textbf{ramen} & \textbf{Mean} & \\
    \midrule
    LLaVA-OV & \textbf{74.87} & 78.93 & 74.08 & 80.86 & 77.19 & 75.22 & 75.44 & 75.33 & 6 \\
    ChatSplat & 74.67 & \textbf{79.17} & \textbf{76.85} & \textbf{80.89} & \textbf{77.90} & \textbf{79.92} & \textbf{76.28} & \textbf{78.10} & \textbf{289} \\
    \bottomrule
    \end{tabular}
    \caption{\textbf{Comparison of Methods on Replica and LERF Datasets with Speed for Object-Level Chatting.} In this comparison, 'OPS' (Objects Per Second) is used as a metric to evaluate the efficiency of each method by measuring the number of objects processed per second. The results clearly demonstrate that ChatSplat is significantly faster than the naive baseline method, highlighting its superior efficiency.}
    \label{tab:object}
\end{table*}
    \section{Additional Results}
        \figref{fig:scene} illustrates the qualitative results of scene-level chatting, showcasing the strengths of ChatSplat. Utilizing multi-view images, ChatSplat achieves a comprehensive understanding of the entire scene, allowing it to generate precise and contextually relevant responses. As demonstrated in \figref{fig:scene}, ChatSplat effectively synthesizes information from multiple perspectives, capturing both spatial and semantic relationships within the scene to facilitate accurate and meaningful conversational interactions.

        The quantitative results for object-level chatting are presented in \tabref{tab:object}. As shown in the table, ChatSplat not only outperforms the baseline in terms of GPT-4o scores but is also significantly faster. The baseline method involves using SAM~\cite{kirillov2023sam} to first mask out an object, followed by LLaVA for chatting with the object. This approach is inherently inefficient, as it requires the integration of two foundation models, both relying on extensive computation for object masking and language feature extraction. In contrast, ChatSplat streamlines the process by rendering object language features for an entire view at once and passing them through an extremely lightweight encoder to generate tokens. This design leads to superior efficiency, enabling ChatSplat to achieve both high answer quality and exceptional processing speed.


\end{document}